\documentclass[a4paper, 10pt, conference]{ieeeconf}      

\IEEEoverridecommandlockouts                              

\usepackage{amsmath} 
\usepackage{amssymb}  
\usepackage{hyperref}
\usepackage{cleveref}
\usepackage[detect-all,per-mode=symbol]{siunitx}
\usepackage{booktabs}
\usepackage{multirow}
\usepackage{pgfplots}
\usepackage[protrusion=true,expansion=true]{microtype} 
\usepackage[keeplastbox]{flushend}
\usepgfplotslibrary{groupplots}
\usetikzlibrary{positioning}
\usepackage[noend]{algcompatible}
\usepackage{algorithm}
\newcommand{\argmax}{\operatornamewithlimits{arg\,max}}

\pgfplotsset{compat=newest}
\pgfplotsset{every axis legend/.append style={%
cells={anchor=west}}
}
\usetikzlibrary{arrows}
\tikzset{>=stealth'}

\usepackage[style=ieee, maxcitenames=2, mincitenames=1]{biblatex}
\AtEveryBibitem{
 	\clearfield{url}  
 	\clearfield{doi}  
\ifentrytype{inproceedings}{
 	\clearlist{address}
 	\clearlist{publisher}
 	\clearname{editor}
 	\clearlist{organization}
 	\clearfield{pages}  
 	\clearlist{location}
 	\clearfield{volume}
 }{}
 }

\AtBeginBibliography{\small}

\usepackage{lipsum}
\usepackage{color}
\usepackage[colorinlistoftodos]{todonotes}

\title{\LARGE \bf
Reinforcement Learning with Iterative Reasoning \\ for Merging in Dense Traffic 
}

\author{Maxime Bouton,$^1$ Alireza Nakhaei,$^2$ David Isele$^3$, Kikuo Fujimura,$^3$ and Mykel J. Kochenderfer$^1$%
    \thanks{*This work was supported by the Honda Research Institute.}
    \thanks{$^{1}$ Maxime Bouton and Mykel J. Kochenderfer are with the Department of Aeronautics and Astronautics, Stanford University, Stanford CA 94305, USA,
            {\tt \{boutonm,mykel\}@stanford.edu}.}%
    \thanks{$^{2}$ Alireza Nakhaei is with the Toyota Research Institute, 4440 El Camino Real, Los Altos, CA 94022, USA, 
            {\tt {alireza.nakhaei}@tri.global}.}%
    \thanks{$^{3}$ David Isele and Kikuo Fujimura are with the Honda Research Institute, 375 Ravendale Dr., Mountain View, CA 94043, USA, 
            {\tt \{disele,kfujimura\}@hra.com}.}%
}

\begin{document}

\maketitle
\thispagestyle{empty}
\pagestyle{empty}

\begin{abstract}
  Maneuvering in dense traffic is a challenging task for autonomous vehicles because it requires reasoning about the stochastic behaviors of many other participants. 
  In addition, the agent must achieve the maneuver within a limited time and distance. 
  In this work, we propose a combination of reinforcement learning and game theory to learn merging behaviors. 
  We design a training curriculum for a reinforcement learning agent using the concept of level-$k$ behavior. 
  This approach exposes the agent to a broad variety of behaviors during training, which promotes learning policies that are robust to model discrepancies.
  We show that our approach learns more efficient policies than traditional training methods.
\end{abstract}

\section{INTRODUCTION}

\begin{figure}[t]
    \centering
    \includegraphics[trim={1cm 9.5cm 0 6.7cm},clip,width=\columnwidth]{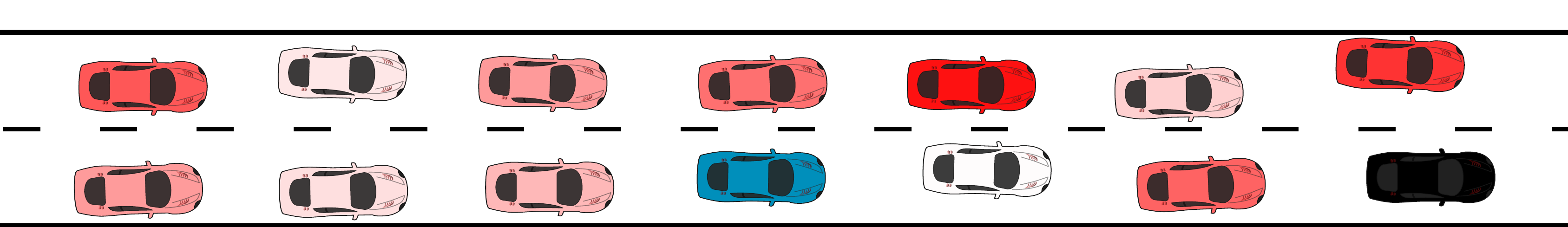}
    \caption{The ego vehicle (in blue) must navigate in very dense traffic. The black vehicle is stopped, forcing the ego vehicle to change lane in a very short distance to move forward. Other vehicles on the road have different behaviors from collaborative (white) to aggressive (dark red).}
    \label{fig:scenario}
\end{figure}

In recent years, major progress has been made to deploy autonomous vehicles and improve safety.
However, certain common driving situations like merging in dense traffic are still challenging for autonomous vehicles. 
Situations like the one illustrated in \cref{fig:scenario} often involve negotiating with human drivers. 
Without good models for interactions with human drivers, standard planning algorithms are often too conservative~\cite{trautman2010}.


Previous attempts to solve this problem have used online search methods and reinforcement learning.
Online search methods rely on computing an action at execution time by sampling different possible scenarios at a finite horizon.
As a consequence, they often have to make coarse assumptions about the driver model to gain computation efficiency. 
\citeauthor{bouton2017} modeled other drivers using Gaussian distributions that do not take into account interaction~\cite{bouton2017}. 
\citeauthor{hubmann2018} used the intelligent driver model (IDM)~\cite{treiber2000} along with a classifier to determine if the ego vehicle is considered the front vehicle. 
These two methods demonstrated promising results but are unlikely to scale to very dense traffic scenarios because the search space grows exponentially with the number of agents~\cite{bouton2017}. \citeauthor{isele2019} proposed a method based on game trees that requires numerous approximations in order to handle dense traffic in realtime, and the approximations can result in suboptimal decisions \cite{isele2019}.
Recently, \citeauthor{bae2020} suggested learning a driver model using a neural network and using that model with an online search method \cite{bae2020}. 
The learned model can lead to good policies but online methods can be computationally expensive during runtime. 

To avoid the computational requirements of online methods, we can use reinforcement learning (RL) instead. 
In RL, the agent interacts with a simulation environment many times prior to execution, and at each simulation episode it improves its strategy. 
The resulting policy can then be deployed online and is often inexpensive to evaluate. 
RL provides a flexible framework to automatically find good decision strategies. 
Recent advances using neural network representations of policies have allowed RL to scale to very complex environments~\cite{mnih2015}.
Such approaches have been successfully applied to autonomous driving applications in lane change scenarios and intersection navigation~\cite{tram2018,hoel2020,saxena2020}. 
However, the scenarios considered often have sparse traffic conditions and consider a small number of other agents, or consider low level action spaces~\cite{saxena2020}.
Additionally, RL agents are known to learn policies that over-fit to the training environment \cite{raghu2018,lanctot2017,mohseni2019}.

In this work, we apply reinforcement learning to navigate dense traffic scenarios (gap of around \SI{2}{\meter} between vehicles). 
A maneuver is a success when the agent passes the stopped vehicles (caused by the black broken car in \cref{fig:scenario}).
In such situation, it requires many decisions for the agent to change lanes and pass the stopped vehicles. 
At any time, a poor decision can lead to a deadlock or an unsafe situation.
These two challenges are commonly referred to as sparse and delayed reward problems.
To address this issue, we propose a curriculum learning approach based on a cognitive hierarchical model to learn efficient policies in challenging environments. 
We use level-$k$ modeling to generate a variety of behaviors in the training environment. 
Each cognitive level is trained in a reinforcement learning environment populated with vehicles of any lower cognitive level.
We show that standard reinforcement learning techniques would fail to learn a good strategy. 
In contrast, our iterative procedure to change the behavior of other agents in the environment allows us to learn robust policies.

\section{BACKGROUND}

This section introduces the notation and fundamental concepts used in this paper. 
We first introduce Markov decision processes and reinforcement learning as our primary approach to address the decision making problem. 
Then we explain the concept of level-$k$ behavior modeling, which inspired the design of our curriculum learning strategy.

\subsection{Reinforcement Learning}

Sequential decision making processes can be modeled as Markov decision processes (MDPs). MDPs are defined by the tuple $(\mathcal{S}, \mathcal{A}, T, R, \gamma)$ where $\mathcal{S}$ is a state space, $\mathcal{A}$ an action space, $T$ a transition model, $R$ a reward function, and $\gamma$ a discount factor.
An agent chooses an action $a\in\mathcal{A}$ in a given state $s$ and receives a reward $r=R(s,a)$. 
The environment then transitions into a state $s'$ according to the distribution $\Pr(s' \mid s, a) = T(s, a, s')$. 

The agent's action is given by a policy $\pi : \mathcal{S} \rightarrow \mathcal{A}$ mapping states to actions.
The agent's goal is to find the policy that maximizes its value, given by the accumulated expected discounted reward given by $\sum_{t=0}^{\infty} \gamma^t r_t$. 
Each policy can be associated to a state-action value function $Q^\pi : \mathcal{S}\times\mathcal{A} \rightarrow \mathbb{R}$, representing the value of following the policy $\pi$.
The optimal state action vaue function of an MDP satisfies the Bellman equation:
\begin{equation}
    Q^*(s, a) = \mathbb{E}_{s'}[R(s, a) + \gamma\max_{a'}Q^*(s', a')]
\end{equation}
The associated optimal policy is given by $\pi^*(s) = \argmax_a Q^*(s,a)$.
In an MDP with large or continuous state spaces, the state action value function can be represented by a parametric model such as a neural network: $Q(s,a; \theta)$.

Reinforcement learning (RL) is a procedure to find the optimal state action value function of an MDP. 
The agent interacts with the environment and gathers experience samples where each sample is in the form of a tuple $(s, a, s', r)$. 
Given an experience sample, the weights of the network can be updated to approximate the Bellman equation as follows:
\begin{equation}
  \theta \leftarrow \theta + \alpha (r + \gamma\max_{a'}Q(s',a';\theta_-) - Q(s,a;\theta))\nabla_\theta Q(s, a;\theta)
\end{equation}
where $\alpha$ is the learning rate and $\theta_-$ represents the weight of a target network.
This algorithm is known as deep Q-learning~\cite{mnih2015}. 
In this work we augment it with prioritized replay~\cite{schaul2016}, dueling~\cite{wang2016}, and double Q-learning~\cite{hasselt2010}.

\subsection{Cognitive Hierarchy Modeling}

Our training curriculum relies on the level-$k$ cognitive hierarchy model from behavioral game theory~\cite{costa1998,wright2010}.
This model consists in assuming that an agent performs a limited number of iterations of strategic reasoning (``I think that you think that I think''). 
A level-$k$ agent acts optimally against the strategy of a level-$(k-1)$ agent~\cite{wright2010}.
Level-$0$ often correspond to a random policy or a heuristic strategy. 
Given a level-$k$ strategy, we can formulate the problem of finding the level-$(k+1)$ policy as the search for an optimal policy in an MDP where the other entities in the environment follow a level-$k$ policy.

In this work, we design a training curriculum to compute the iterative best response up to a given level of strategic reasoning. 
Given a strategic level $k$, we can solve $k$ RL problems iteratively to compute a strategy of level $k$ by populating the training environment with agents of lower strategic levels.
Such iterative procedure is not guaranteed to converge as $k$ goes to infinity. However, if it does, the solution is a Nash equilibrium~\cite{nash1950, dmu}.
In practice, we only consider a limited number of iterations. 
Previous work in behavioral game theory showed that a reasoning level of \num{2} or \num{3} is a better approximation to human behavior than Nash equilibrium~\cite{wright2010,dmu}.
It is then reasonable to limit the reasoning level of an autonomous agent to \num{4} to anticipate human behaviors.

\section{AGENT DESIGN}

This section describes the design of our RL agent. 
We first explain the design of a high level action space as well as the underlying dynamics model used for training. 
Then, we describe the input features and the network architecture.

\subsection{Action Space}
\label{sec:actionspace}

\begin{figure}[t]
    \centering 
    \includegraphics[width=0.5\columnwidth]{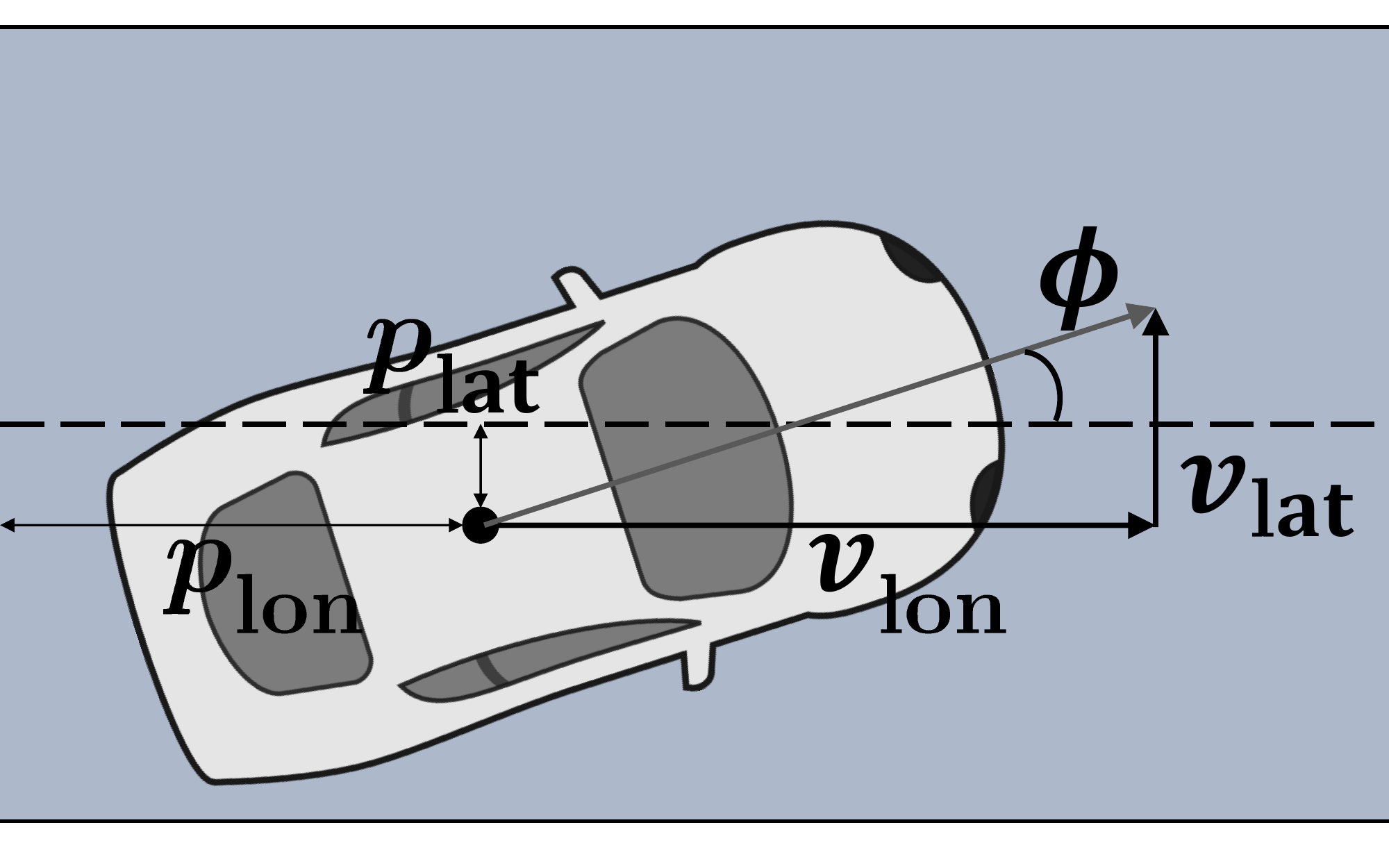}
    \caption{Description of the state of the agent. The longitudinal and lateral position are described with respect to the center line of its lane (dashed line). The longitudinal and lateral velocities are function of the orientation of the vehicle with respect to the center line.}
    \label{fig:state}
\end{figure}

The state of the agent is described by four quantities: longitudinal and lateral positions and longitudinal and lateral velocities as illustrated in \cref{fig:state}.
The state of other entities is described relative to the ego vehicle state.
Our agent controls both its longitudinal and lateral motions.
The longitudinal motion is controlled using the intelligent driver model~\cite{treiber2000}. 
The agent chooses among three desired speed levels: \SI{0}{\meter\per\second}, \SI{3}{\meter\per\second}, \SI{5}{\meter\per\second}.
This decision is converted into a longitudinal acceleration using the equation from the IDM model~\cite{treiber2000}.
By using the IDM rule to compute the acceleration, the behavior of braking if there is a car in front will not have to be learned. 
The longitudinal action space is safe by design. 
This can be thought of as a form of shield to the RL agent from taking unsafe actions \cite{alshiekh2018,isele2018}. 
Other drivers have stochastic behaviors and might still caused collisions.

The lateral motion is controlled by two actions: stay or change lane. 
A lateral acceleration is then computed using a proportional derivative controller:
\begin{equation}
  a_\text{lat} = -k_p p_\text{lat} - k_d v_\text{lat} 
  \label{eq:pdcontrol}
\end{equation}
where $p_\text{lat}$ is the lateral position of the vehicle with respect to the center line of the lane and $v_\text{lat}$ is the lateral velocity of the vehicle, in the direction normal to the center line, $k_p$ and $k_d$ are controller gains (set to \SI[per-mode=reciprocal]{3}{\per\second\squared} and \SI[per-mode=reciprocal]{3}{\per\second} respectively).
Given the dense nature of the traffic, we did not add any safety rule when moving laterally. 
Enforcing a minimum gap size would cause the agent to fail the maneuver because large gaps are rare in our scenario.

There is a total of six actions: the Cartesian product of the three longitudinal actions and two lateral actions. 
The lateral and longitudinal acceleration commands are used to update the physical state of the agent using the following dynamics: 
\begin{align}
  p'_\text{lon} &= p_\text{lon} + v_\text{lon} \delta t \\
  p'_\text{lat} &= p_\text{lat} + v_\text{lat} \delta t \\
  v'_\text{lon} &= v_\text{lon} + a_\text{lon}\delta t \\
  v'_\text{lat} &= v_\text{lat} + a_\text{lat}\delta t
\end{align}
where $\delta t$ is a simulation step (\SI{0.1}{\second}), and the primed quantities correspond to the updated value of the state.
This simple dynamics model is fast to simulate but not accurate. 
To provide robustness to model inaccuracies, we added constraints on the steering rate and the maximum steering angle of \SI{0.4}{\radian\per\second} and \SI{0.5}{\radian}, respectively. 
These constraints eventually limit the lateral motion of the vehicle.

The agent takes an action every five simulation steps (\SI{0.5}{\second} between two actions).
The learned policy is intended to be high level. At deployment, we expect the agent to decide on a desired speed and a lane change command while a lower lever controller, operating at higher frequency, is responsible for executing the motion and triggering emergency braking system if needed.

\subsection{Input Features}

Each vehicle in the environment is represented by its longitudinal and lateral position, longitudinal and lateral velocity, a heading angle, as well as a longitudinal and lateral acceleration. 
We assume that the agent can observe its own state perfectly. 
In addition the agent can observe vehicles in the neighboring lanes within \SI{30}{\meter}. For each vehicle present in this field of view, the ego vehicle can measure:
\begin{itemize}
    \item Relative longitudinal and lateral position.
    \item Relative longitudinal and lateral velocity.
\end{itemize}

As the number of vehicles within the field of view of the agent varies, we cap this number to the eight closest vehicles to provide a fixed size input to our RL agent. If there are fewer than eight agents in the vicinity of the ego vehicle, we send a fixed feature vector associated to absent vehicles. 

This work assumes perfect measurements. 
However, the agent does not have full observability of the environment because it can neither observe the acceleration of other agent nor the internal states governing their behavior. 
Measurement uncertainty can be handled online (after training) using the QMDP approximation technique~\cite{bouton2019}.

\subsection{Policy Network}

\begin{figure}
    \centering
    \includegraphics[width=\columnwidth]{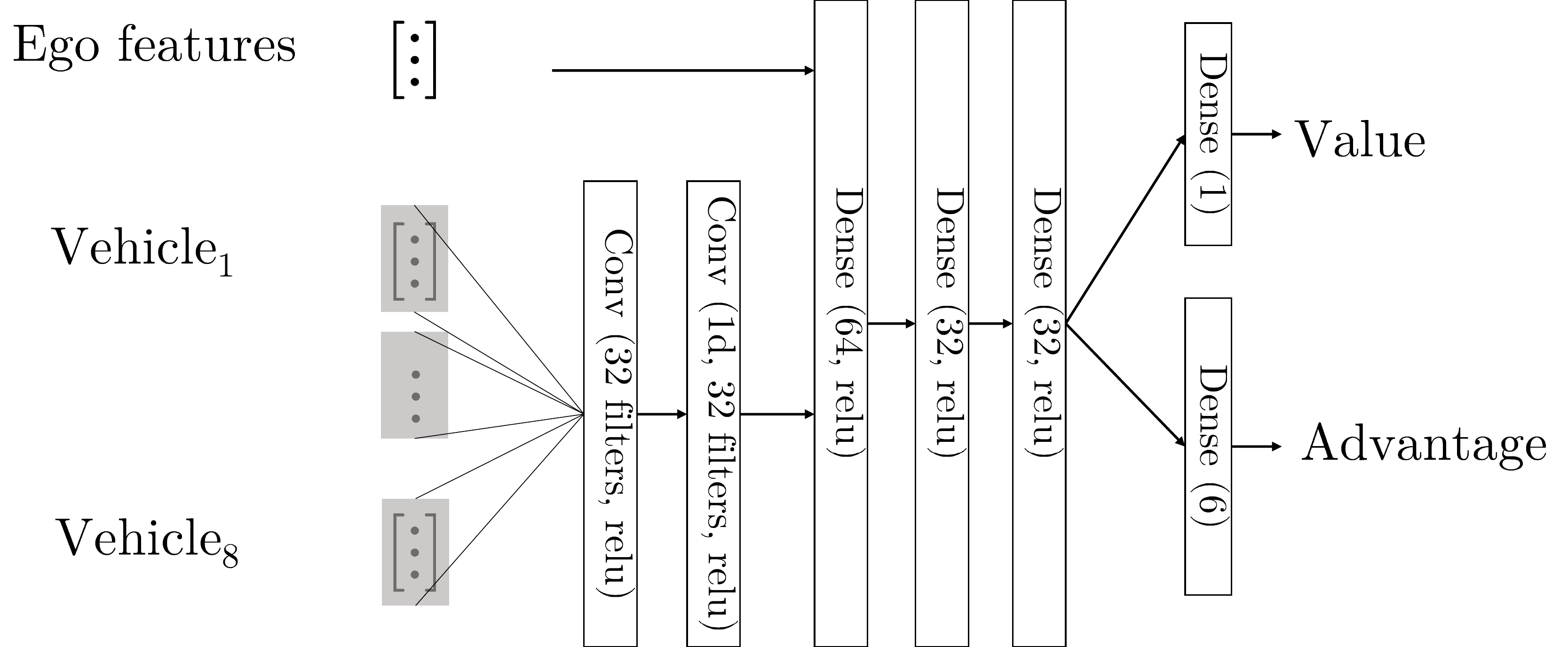}
    \caption{Illustration of the network architecture~\cite{hoel2020}. The convolutional layers allow to share the weights when processing other vehicles' states.}
    \label{fig:network_architecture}
\end{figure}

We use a network architecture similar to \citeauthor{hoel2020}~\cite{hoel2020}. 
Our early experiments showed that training was faster and more stable using this architecture compared to standard fully connected networks.
The architecture used is depicted in \cref{fig:network_architecture}.
The input to the network is divided into two sets of features: the ego features, and the other vehicles features (positions and speeds). 
The other vehicle features are processed using a convolutional network which ensures translation independence between the features. 
The network is learning a similar representation for position, and velocity independently of the associated vehicle. 
The other vehicle features are then combined with the ego vehicle features using a fully connected layer. 
The policy network maps the input features to two streams, a value stream representing the value of the current observation, and an advantage stream representing the advantage of each action. 
The agent chooses the action with the best advantage. 
Splitting the end of the policy network between advantage and value is called dueling and is commonly used to improve standard DQN~\cite{wang2016}.

This section described the design of our RL agent. 
The next section describes how to design an efficient training procedure to avoid standard issues with RL: sparse rewards, delayed rewards, and generalization.

\section{ITERATIVE REASONING CURRICULUM}

This section describes the main contribution of this paper, which is using level-$k$ modeling to design an efficient training curriculum.

\subsection{Overview of the Approach}

Many reinforcement learning researchers use predefined games as benchmarks for evaluating algorithms. 
When applying those algorithms to other problems, such as autonomous driving, designing the training environment is often challenging.
The model of the agent, as well as the behavior of traffic participants in the simulation environment should match the real world as closely as possible. 
In addition, the environment must be sufficiently diverse such that the agent can generalize to a variety of situations.
The training curriculum proposed in this work allows learning a variety of behaviors, and diversifies the population of drivers in the environment as the cognitive level increases. 
An agent with a high level of reasoning will then have been exposed to a variety of behaviors, and, as a result, its policy should generalize better.

The proposed curriculum learning is based on the level-$k$ behavior model and is illustrated in \cref{alg:curriculum}.
Initially, an agent is trained to perform a lane change in crowded traffic where all other agent follow a level-$0$ policy. 
The level-$0$ policy is a hand-engineered rule-based policy defined in the next section. 
The trained agent is then referred to as level-$1$. 
The next step of the curriculum is to train an agent to follow the top lane safely, while agents on the bottom lane follow either a level-$0$ or level-$1$ policy. 
This agent is referred to as level-$2$. 
We then train a level-$3$ agent by populating the top lane with level-$0$ and level-$2$ agents and the bottom lane with  level-$0$ or level-$1$ agents. 
The procedure is then repeated until a sufficiently high level of reasoning is reached. 
The agents with an odd reasoning level are performing lane change maneuvers (merging agents) while the agents with an even reasoning level are performing a keep lane maneuver.
The level-$0$ agents can perform both maneuvers and are spread between the two lanes. Note that unlike traditional level-$k$ modeling, where a level-$k$ agent optimizes against a level-$(k-1)$ agent, our level-$k$ agent optimizes against the population of level-$0$ through level-$(k-1)$ agents.

\begin{algorithm}
	\caption{Curriculum with Iterative Reasoning}
	\begin{algorithmic}[1]
		\STATE \textbf{input: } level-$0$ policy, maximum reasoning level $N$
		\FOR{$k=1\ldots N$}
		    \STATE sample reasoning levels from $\text{Uniform}(0, k-1)$
		    \STATE populate bottom lane with merging agents
		    \STATE populate top lane with keep-lane agents
            \STATE train level-$k$ agent using deep Q-learning
		\ENDFOR
	\end{algorithmic}
    \label{alg:curriculum}
\end{algorithm}

The curriculum's main parameters are the maximum reasoning level $N$, the level-$0$ policy, and the distribution over the reasoning level. 
The maximum reasoning level must be carefully chosen. 
At each step of the curriculum one can evaluate the performance of the new level and decides whether to continue the training or not. 
In this work we trained policies up to level \num{5}.
To accelerate training at each time step, we re-use the weights from the previous iteration to start training. 
Since the even agents and odd agents are learning different tasks (merge or keep lane), we make sure to initialize the weight with the previous level corresponding to the same task.
Note that a level 1 policy corresponds to a standard RL procedure.

\subsection{Level-$0$}\label{sec:level-0}

The level-$0$ policy consists of a combination of a longitudinal driver model, and a lane change model. The longitudinal model has a constant desired speed drawn from a distribution, and the lateral model chooses to change lane or not based on a set of hand-engineered rules. Similarly as for our agent design, these decisions are converted into a longitudinal acceleration $a_\text{lon}$ and a lateral acceleration $a_\text{lat}$.

The longitudinal model is an extension of the IDM model with a cooperation parameter and a perception parameter inspired by previous work~\cite{isele2019, bouton2019cooperation, bae2020}.
A parameter $\eta_\text{percept}$ determines a yield area, if a vehicle is in this yield area, a yield action is sampled according to a Bernoulli distribution with parameter $c$. 
If $c=1$, the driver always yields.
This longitudinal driver model allows to compute a longitudinal acceleration: $a_\text{lon}$. 

The lateral motion is governed by a lane tracker and a lane change model. 
We use MOBIL~\cite{kesting2007} to decide when to perform the lane change. 
Given a desired lane, the lateral acceleration $a_\text{lat}$ is given by the PD controller described in \cref{eq:pdcontrol}. 
The state of the vehicle is then updated using the dynamics model described in \cref{sec:actionspace}. 

To make the training environment diverse, we sampled different level-$0$ agents by changing the parameters of the longitudinal model. 
The distribution of the parameters is given in \cref{tab:driverdist}. 
Adding noise in the driver model parameters helps the higher level agents generalize, as they will be exposed to a variety of behavior during training.

\begin{table}[h]
	\centering
	\caption{Cooperative IDM distribution}
	\begin{tabular}{@{}ll@{}}
		\toprule[1pt]
		Parameter &  Distribution \\
		\midrule
    $\eta_\text{percept}$ & $\text{Uniform}(-0.15, 0.15)$ \\
    $c$ & $ \text{Uniform}(0, 1)$ \\
    $\delta$ & $\text{Uniform}(3.5, 4.5)$ \\
    $T$ & $\text{Uniform}(3.5, 4.5)$ \SI{}{\second} \\
    $s_\text{min}$  & $\text{Uniform}(1.0, 2.0)$ \SI{}{\meter} \\
    $a_\text{max}$ & $\text{Uniform}(2.5, 3.5)$ \SI{}{\meter\per\second\squared}\\
    $d_\text{cmf}$ & $\text{Uniform}(1.5, 2.5)$ \SI{}{\meter}\\
    $v_\text{des}$ & $\text{Uniform}(2.0, 5.0)$ \SI{}{\meter\per\second}\\
		\bottomrule[1pt]
	\end{tabular}
	\label{tab:driverdist}
\end{table}

\subsection{Reward function}

Contrary to standard level-$k$ reasoning, the even levels and odd levels correspond to different tasks (keep lane or change lane). 
For both tasks, the reward function has the same additive structure. 
The reward function consists of the following terms: 
\begin{itemize}
    \item Penalty for collisions: \num{-1}
    \item Penalty for deviating from a desired velocity: ${-0.001 |v_\text{ego} - v_\text{desired}|}$
    \item Reward for being in the top lane: +\num{0.01} for the merging agent and \num{0} for the keep lane agent.
    \item Reward for passing the blocked vehicle: +\num{1}
\end{itemize}

The weight of each component is designed to keep the accumulated reward within a reasonable numerical range to help the convergence of the Q-network.


\section{RESULTS}

\subsection{Experiments}

The training environment models both dense and slow traffic corresponding to a traffic jam in an urban area. 
The gap between vehicles varies between \SI{1}{\meter} and \SI{6}{\meter}. 
When initializing a training episode, each car has a random speed, random lateral position and random heading given by the distributions in \cref{tab:initial-scene}.
To have the policies generalize to different traffic conditions, we sample the number of cars uniformly from \num{10} cars (sparse traffic) to \num{50} cars (dense traffic). 
Evaluation is done in dense traffic only, where the number of cars is always \num{50}. The simulation environment and the algorithm are implemented using the Julia language~\cite{bezanson2017julia}. Each level takes an hour to train on a single CPU.

\begin{table}[h]
	\centering
	\caption{Initial Scene Parameters}
	\begin{tabular}{@{}ll@{}}
		\toprule[1pt]
		Parameter &  Value \\
		\midrule
    $v_0$ & $\text{Uniform}(1.0, 2.0)$  \SI{}{\meter\per\second}\\
    $t_0$ & $\text{Uniform}(-0.75, 0.75)$  \SI{}{\meter}\\
    $\phi_0$ & $\text{Uniform}(-0.1, 0.1)$  \SI{}{\radian}\\
    $n_\text{cars}$ & \text{Uniform}(10, 50) \\
    gap & \SI{6}{\meter} \\
		\bottomrule[1pt]
	\end{tabular}
	\label{tab:initial-scene}
\end{table}

A scenario has three possible outcomes: a success, a collision, or a time-out. 
A success is declared when the ego vehicle merges onto the desired lane and stays there for \SI{5}{\second}. 
This success criteria imposes the vehicle to merge safely since it must not cause any crash in a short time period after the merge. 
A time-out is declared if the ego vehicle stays in the initial lane for more than \SI{40}{\second}.

\begin{figure}
    \centering
    \input{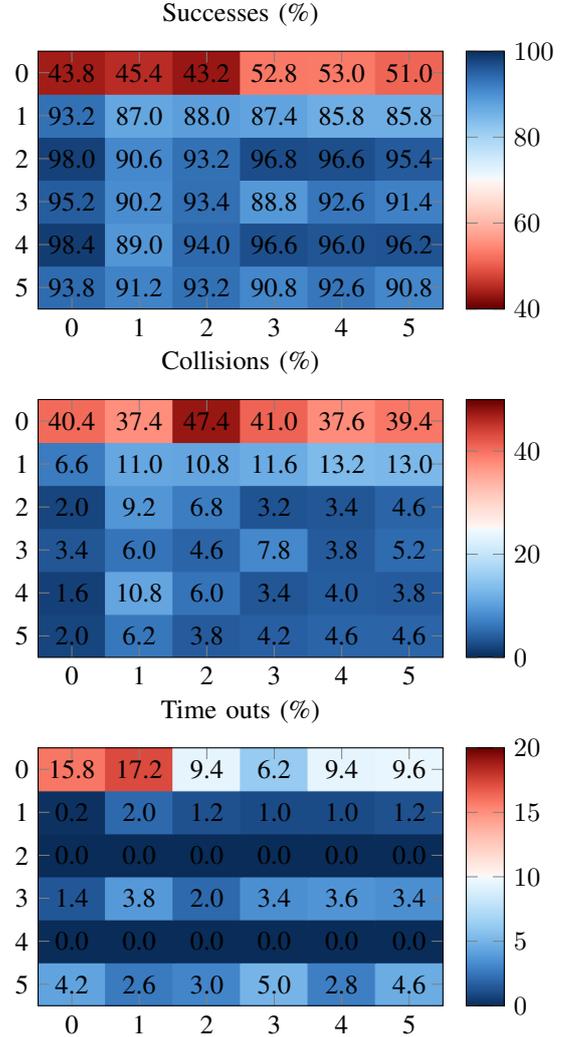}
    \caption{We evaluated each policy in environments with agents of varying reasoning levels. In a level $k$ environment, an agent can encounter any agents of level smaller or equal to $k$. The horizontal axis represents the level of the environment, and the vertical axis represents the level of the agent being evaluated.}
    \label{fig:matrix-plot}
\end{figure}

We trained five different policies according to the procedure described in \cref{alg:curriculum}. The performance of those policies is summarized in \cref{fig:matrix-plot} where each cell is the result of \num{500} simulations.
Policies \num{1}, \num{3}, and \num{5} correspond to agents that learned to change lane, and policies \num{2} and \num{4} correspond to keep lane agents. 
The level \num{0} policy is a hand engineered policy described in \cref{sec:level-0}. 
The rows in \cref{fig:matrix-plot} represents the policy being evaluated, the number corresponds to the reasoning level. 
The columns correspond to the level of other vehicle in the environment. 
A level \num{5} environment has \num{50} other traffic participants of reasoning levels ranging from \num{0} to \num{5}. 
\cref{fig:matrix-plot} shows the effect of the reasoning level of the agent on the three different metrics and its ability to generalize to unseen behavior. 
When evaluating a level \num{3} policy against a level \num{5} environment, the agent faces behaviors that have not been seen during training (levels \num{3}, \num{4}, and \num{5}). 

\begin{figure}
    \centering
    \input{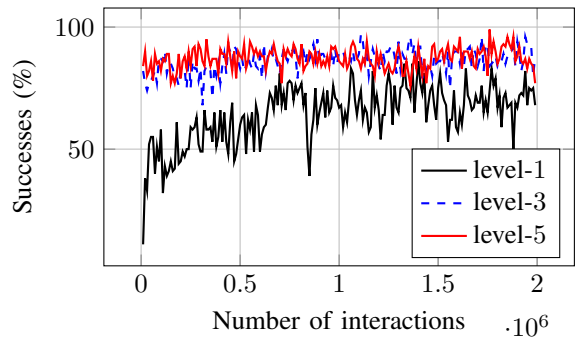}
    \caption{Training performance of three policies at different reasoning level in the curriculum. The percentage of successes between level \num{3} and \num{5} is very similar which indicates that higher reasoning levels might not be needed for higher performance.}
    \label{fig:training}
\end{figure}

Our second experiment analyzes the convergence of our training algorithm, and the convergence of the curriculum based training. 
In \cref{fig:training}, we show the evolution of the success rate with the number of interaction with the environment. 
We compare the learning curves of three different levels (\num{1}, \num{3}, and \num{5}), the three agents learning to merge. 
At each level we use the policy from the previous level to start training. 
As a consequence, the policy is already performing efficiently even with only a small number of interactions.

\subsection{Discussion}

In the matrix plots from \cref{fig:matrix-plot}, we can see that the hand-engineered policy does not perform well in this scenario. 
The lane change model MOBIL~\cite{treiber2000,kesting2007}, which is at the core of this rule-based policy has been designed for sparse traffic conditions. 
Although we tuned the parameters, it is still not appropriate to navigate such dense and stochastic traffic. 

Levels \num{1}, \num{3}, and \num{5} learned efficient policies with more than \SI{90}{\percent} success regardless of the environment. 
Contrary to level \num{1}, we can see that level \num{3} and \num{5} generalize better as they have a better performance against the higher level environments, even those which were not seen during training. None of the learned policies were able to learn a safe behavior as the collision rate is higher than \SI{2}{\percent}. 
However, the learned policies are safer than the rule-based method and have fewer time outs. 
The remaining collisions could be addressed as future work by using an emergency braking system. 
A more thorough reward engineering could also help reduced that number. 
The focus of this paper is to show the benefit of the level-$k$ curriculum approach for training RL agents more efficiently in complex traffic scenarios.

Our approach requires to choose an appropriate reasoning level. 
The results from \cref{fig:matrix-plot} and \cref{fig:training} show the relative improvements of successive reasoning levels.
Both figures demonstrate that there is little performance improvement between level \num{3} and \num{5}, which indicates that larger reasoning levels might not be needed for this task.

\section{CONCLUSIONS}

This paper presented a reinforcement learning curriculum based on level-$k$ reasoning to learn to merge in dense traffic. 
We showed how to design a high level decision making agent and a training environment for the merging scenario. 
Then we demonstrated a curriculum learning with iterative reasoning. 
At each step of the curriculum, the behavior of the entities in the environment is sampled from the previous levels of the curriculum. 
As the reasoning level increases, the learning agent is exposed to a larger variety of behaviors.
We demonstrated empirically the benefit of this approach for learning policies that generalize to behaviors unseen during training. 
By iteratively increasing the reasoning level, we are able to learn a more efficient and more general policy than with standard reinforcement learning techniques.

Future work involves combining this policy with a low level controller and transferring it to the real world. 
We also plan to investigate using the learned policy with online planning methods. 
The policy could be used to compute the cost to go in an MPC planner~\cite{bae2020}, which could enforce stronger safety constraints than the reinforcement learning agent.


\addtolength{\textheight}{-12cm}   


\printbibliography





\end{document}